\documentclass[runningheads]{llncs}

\usepackage{times}
\usepackage{epsfig}
\usepackage{bbm}
\usepackage{array}
\usepackage{verbatim}
\usepackage{textcomp}
\usepackage[font=small,skip=0pt]{caption}
\usepackage{graphicx}
\usepackage{amsmath,amssymb} 
\usepackage{color}
\usepackage{mathtools}
\usepackage{subfig}
\usepackage[width=122mm,left=12mm,paperwidth=146mm,height=193mm,top=12mm,paperheight=217mm]{geometry}

\begin{document}
\pagestyle{headings}
\mainmatter
\title{MKL-RT: Multiple Kernel Learning for Ratio-trace Problems via Convex Optimization} 

\titlerunning{Multiple Kernel Learning for Ratio-trace Problems}

\authorrunning{Raviteja Vemulapalli, Vinay Praneeth Boda, Rama Chellappa}

\author{Raviteja Vemulapalli, Vinay Praneeth Boda, Rama Chellappa}
\institute{University of Maryland, College Park}

\maketitle

\begin{abstract}
In the recent past, automatic selection or combination of kernels (or features) based on multiple kernel learning (MKL) approaches has been receiving significant attention from various research communities. Though MKL has been extensively studied in the context of support vector machines (SVM), it is relatively less explored for ratio-trace problems. In this paper, we show that MKL can be formulated as a convex optimization problem for a general class of ratio-trace problems that encompasses many popular algorithms used in various computer vision applications. We also provide an optimization procedure that is guaranteed to converge to the global optimum of the proposed optimization problem. We experimentally demonstrate that the proposed MKL approach, which we refer to as MKL-RT, can be successfully used to select features for discriminative dimensionality reduction and cross-modal retrieval. We also show that the proposed convex MKL-RT approach performs better than the recently proposed non-convex MKL-DR approach.
\end{abstract}
\section{Introduction}
In many computer vision applications, we are often interested in transforming our initial feature representation $x$ to a new representation $x^{\prime}$ such that $x^{\prime}$ suits better for the application under consideration. For example, if we are interested in classification, we may want to transform $x$ such that samples from the same class are close to each other and samples from different classes are far away from each other. If we are interested in retrieving images using text query, we may want to transform our image representation such that it is highly correlated with the corresponding text representation.

If we plan to use a linear transformation, then we are interested in learning a transformation matrix $W$, such that $x^{\prime} = W^\top x$ has certain desired properties depending on the application of interest. Though various different algorithms (aiming at different applications) have been proposed in the past for learning the transformation matrix $W$, many of them end up solving a ratio-trace problem~\cite{Yan07,Wang07} (equation~\eqref{eqn::ratio-trace-lin}), whose optimal solution can be obtained using generalized Eigenvalue decomposition (GEVD)~\cite{Fukunaga91}.

Algorithms based on ratio-trace problems have been extensively used in various computer vision applications~\cite{Yan07,Etemad97,Belhumeur97,Sharma12,Cai07,Meina11,Chen05,Kim07,Rasiwasia10}. Some of the popular algorithms formulated as a ratio-trace problem (equation~\eqref{eqn::ratio-trace-lin}) are linear discriminant analysis (LDA)~\cite{Fukunaga91}, semi-supervised discriminant analysis (SDA)~\cite{Cai07}, side-information based LDA (SILDA)~\cite{Meina11}, local discriminant embedding (LDE)~\cite{Chen05}, marginal Fisher analysis (MFA)~\cite{Yan07}, locality preserving projections (LPP)~\cite{Hen03}, neighborhood preserving embedding (NPE)~\cite{He05}, canonical correlation analysis (CCA)~\cite{Hardoon04}, and orthonormal PLS-SB~\cite{Rosipal05}.

All the above mentioned linear algorithms suffer from two main disadvantages: (i) They require input data to be represented in the form of feature vectors $x$ in an Euclidean space. This may not be possible in applications where the data of interest is represented using bag-of-features~\cite{Lazebnik06}, matrices~\cite{Tuzel06} or manifold features~\cite{Vemulapalli13}. In some applications, we may only have similarities or distances between the features instead of explicit representations. (ii) Linear transformations may be too simple to be effective in some applications as they can not handle non-linearity present in the data. Both of these issues can be handled by using kernels. Kernelized versions of these linear algorithms also end up solving a ratio-trace problem (equation~\eqref{eqn::ratio-trace-ker}).

Though kernel-based methods have been successfully used in many computer vision applications, the kernel function and the associated feature space are central choices that are generally made by the user. Recently, automatic selection or combination of kernels (or features) based on MKL approaches has been shown to produce state-of-the-art results in various applications~\cite{Vedaldi09,Gehler09,Yeh12}. Multiple kernel learning was initially proposed~\cite{Lanckriet03} for SVM and has since received significant attention~\cite{Sonnenburg06,Rakotomamonjy08,Kloft11}. An excellent overview of various MKL algorithms can be found in~\cite{Gonen11}.

Though MKL has been extensively studied in the context of SVM, it is relatively less explored for ratio-trace problems. Motivated by MKL-SVM, Kim~\textit{et.\ al.}~\cite{Kim06} and Ye~\textit{et.\ al.}~\cite{Ye08} extended the MKL approach to LDA (which is a specific instance of ratio-trace problem~\eqref{eqn::ratio-trace-lin}) formulating it as a convex optimization problem. Arguing for non-sparse MKL, Yan~\textit{et.\ al.}~\cite{Yan12} proposed a non-sparse version of MKL-LDA, which imposes a general $\ell_p$ norm regularization on the kernel weights. 

Motivated by MKL-LDA, Lin~\textit{et.\ al.}~\cite{Lin11} extended the MKL approach to graph embedding framework~\cite{Yan07} which covers a large number of dimensionality reduction algorithms. The MKL-DR framework of~\cite{Lin11} is based on trace-ratio formulation which is different from the ratio-trace formulation used in this paper. We refer the readers to~\cite{Wang07} for a discussion on the differences between trace-ratio and ratio-trace formulations. The trace-ratio based MKL-DR formulation of~\cite{Lin11} results in a non-convex optimization problem. In~\cite{Lin11}, the authors used an iterative optimization procedure that has no convergence guarantees.

In this paper, we show that similar to MKL-LDA~\cite{Ye08}, kernel learning can be formulated as a convex optimization problem for a large class of ratio-trace problems (equations~\eqref{eqn::ratio-trace-lin} and~\eqref{eqn::ratio-trace-ker}) that includes popular algorithms like LDA, SDA, SILDA, LDE, NPE, MFA, LPP, CCA and orthonormal PLS-SB. We also provide an optimization procedure that is guaranteed to converge to the global optimum of the proposed convex optimization problem.

In practice, MKL is typically used in two different ways: (i) Various kernels are defined for the same feature representation, for example Gaussian kernels with different values of $\sigma$, and an optimal kernel is learned as a combination of these kernels~\cite{Lanckriet03,Ye08}, (ii) Multiple feature descriptors are used to represent objects of interest and a similarity kernel is generated from each feature~\cite{Gehler09,Lin11}. In this case, kernel learning effectively solves the feature selection problem and the MKL coefficients can be interpreted as weights given to the corresponding features. One can also use a mixed strategy~\cite{Yeh12} of using multiple features and defining multiple kernels for each feature. In this paper, we use MKL for feature selection in the context of ratio-trace problems.

\textbf{Contributions:} 1) We show that MKL can be formulated as a convex optimization problem for a large class of ratio-trace problems. The proposed MKL-RT formulation is applicable to various popular algorithms like LDA, SDA, SILDA, LDE, NPE, MFA, LPP, CCA and orthonormal PLS-SB. 2) We provide an optimization procedure that is guaranteed to converge to the global optimum of the proposed optimization problem.\hspace{4pt} 3) We experimentally show that the proposed MKL-RT approach can be successfully used to select features for discriminative dimensionality reduction and cross-modal retrieval. 4) We show that the proposed ratio-trace based convex MKL-RT approach performs better than the trace-ratio based non-convex MKL-DR approach of~\cite{Lin11}.

\textbf{Organization:} Section 2 presents the general class of ratio-trace problems for which MKL can be formulated as a convex optimization problem. Section 3 discusses some specific instances of the ratio-trace problem which will be used in our experimental evaluation. Section 4 presents the proposed convex MKL-RT formulation. We present our experimental results in section 5 and conclude the paper in section 6. 

\textbf{Notations:} We use $\mathbbm{1}$ to denote the indicator function. The transpose of a matrix $A$ is denoted by $A^\top$. We use $I$ to denote an identity matrix of appropriate size. We use $|\cdot|$ to denote the absolute value and $\emptyset$ to denote the empty set.
\section{Ratio-trace Problem}
\textbf{Definition 1:} For any two $d \times d$ symmetric positive semi-definite matrices $S_1$ and $S_2$, the ratio-trace problem is defined as 
\begin{equation}
\underset{W}{\text{maximize}}\ \ \ \text{trace}\left[\left(W^\top S_1 W \right)^{-1} \left(W^\top S_2 W\right)\right].
\label{eqn::ratio-trace}
\end{equation}

In this paper, we focus on the class of algorithms that learn the data transformation matrix $W$ by solving the following ratio-trace problem:
\begin{equation}
\underset{W}{\text{maximize}}\ \ \ \text{trace}\left[\left(W^\top \left((1-\sigma)X L X^\top + \sigma I\right)W \right)^{-1} \left(W^\top X L^{\prime} X^\top W\right)\right],
\label{eqn::ratio-trace-lin}
\end{equation}
where $X = [x_1, \hdots, x_N] \in \mathcal{R}^{d\times N}$ is the data matrix (assumed to be centered), $\sigma \in (0, 1)$ is a regularization parameter used to prevent overfitting, and $L (\neq 0),\ L^{\prime}$ are (algorithm-dependent) $N \times N$ symmetric positive semi-definite matrices. 

Some of the popular algorithms that fall into this class are LDA, SDA, SILDA, LDE, MFA, LPP, NPE, CCA and orthonormal PLS-SB. All these linear algorithms can be made non-linear by using kernels. For a given kernel function $\mathcal{K}$, the kernelized versions of these algorithms solve the following ratio-trace problem:
\begin{equation}
\underset{\Gamma}{\text{maximize}}\ \ \ \text{trace}\left[\left(\Gamma^\top \left((1-\sigma)K L K + \sigma K\right)\Gamma\right)^{-1}\left(\Gamma^\top K L^{\prime} K \Gamma\right)\right],
\label{eqn::ratio-trace-ker}
\end{equation}
where $K$ is the $N \times N$ kernel matrix with $K_{ij} = \mathcal{K}(x_i, x_j)$. The optimal solution to~\eqref{eqn::ratio-trace-ker} is given by the generalized Eigenvectors~\cite{Fukunaga91} corresponding to the non-zero generalized Eigenvalues of the matrix pair $\left(KL^{\prime}K,\ (1-\sigma)KLK + \sigma K\right)$ :
\begin{equation}
KL^{\prime}K \gamma = \lambda \left((1-\sigma)KLK + \sigma K\right)\gamma.
\label{eqn::gevd}
\end{equation}
Once $\Gamma$ is obtained, the new (non-linearly transformed) representation for a data sample $x \in \mathcal{R}^d$ can be computed using $x^{\prime} = \Gamma^\top [\mathcal{K}(x_1, x),\hdots, \mathcal{K}(x_N, x)]^\top.$
\section{Instances of the Ratio-trace Problem}
\label{sec::Instances}
As mentioned earlier, various algorithms~\cite{Yan07,Cai07,Meina11,Chen05,Hen03,He05,Hardoon04,Rosipal05} used in many computer vision applications are formulated as a ratio-trace problem. In this section, we briefly discuss three specific instances of the ratio-trace problem which will be used in the experimental evaluation of the proposed MKL-RT approach in section~\ref{sec::Experiments}.
\subsection{KFDA - Kernel Fisher Discriminant Analysis}
KFDA~\cite{Baudat00} is a popularly-used non-linear discriminative dimensionality reduction algorithm. Let $\{(x_i, y_i),\ i = 1,\hdots,N\}$ be the set of $N$ labeled training samples, where $y_i \in \{1,\hdots,P\}$ is the class label of feature $x_i \in \mathcal{R}^d$. Let $y = [y_1,\hdots, y_{N}]$, $\vec{1}_p = \mathbbm{1}[y = p]$ be the membership vector corresponding to class $p$, and $N_p$ be the number of labeled samples in class $p$. Let $\mathcal{K}$ be a kernel function defined on features $x_i$ and $K^x$ be the corresponding kernel matrix. Then, KFDA solves the ratio-trace problem~\eqref{eqn::ratio-trace-ker} with
\begin{equation}
K = K^x,\ \ \ L^{\prime} = \sum_{p=1}^P \frac{1}{N_p}\vec{1}_p \vec{1}_p^\top,\ \ \text{and}\ \  L = I - L^{\prime}\ \text{or}\ L = I.
\end{equation}
The lower dimensional representation for a data sample $x \in \mathcal{R}^d$ can be computed using $x^{\prime} = \Gamma^\top [\mathcal{K}(x_1, x), \hdots, \mathcal{K}(x_N, x)]^\top.$
\subsection{KCCA - Kernel Canonical Correlation Analysis}
KCCA~\cite{Hardoon04} is a popular approach used in cross-modal retrieval applications. KCCA maps the data (non-linearly) from two different modalities to a common lower dimensional latent/concept space where the two modalities are highly correlated. Let $\{(x_i, z_i),\ i = 1,\hdots,N\}$ be $N$ training data pairs where $x_i \in \mathcal{R}^{d_1}$ and $z_i \in \mathcal{R}^{d_2}$ are samples from the first and second modalities respectively. Let $\mathcal{K}^x$ and $\mathcal{K}^z$ be kernel functions defined on features $x_i$ and $z_i$ respectively. Let $K^x$ and $K^z$ be the corresponding kernel matrices. Then, KCCA solves the ratio-trace problem~\eqref{eqn::ratio-trace-ker} with
\begin{equation}
K = K^x,\ \ \ L^{\prime} = K^z\left((1-\sigma)K^z + \sigma I\right)^{-1}\ \ \text{and}\ \ L = I .
\end{equation}
The latent space representations for samples $x \in \mathcal{R}^{d_1}$ and $z \in \mathcal{R}^{d_2}$ from first and second modalities respectively are given by  $x^{\prime} = \Gamma^\top [\mathcal{K}^x(x_1, x),\hdots, \mathcal{K}^x(x_N, x)]^\top$ and $z^{\prime} = \Xi^\top [\mathcal{K}^z(z_1, z), \hdots, \mathcal{K}^z(z_N, z)]^\top$, where 
\begin{equation}
\Xi = ((1-\sigma)K^z + \sigma I)^{-1}K^x \Gamma\Lambda^{-\frac{1}{2}}.
\end{equation}
Here, $\Lambda$ is the diagonal matrix of non-zero generalized Eigenvalues given by~\eqref{eqn::gevd}.
\subsection{LKCCA - Labeled Kernel Canonical Correlation Analysis}
KCCA requires paired training data samples $\{(x_i, z_i)\}$ from two modalities to learn the transformations from the initial feature spaces to the common latent space. Suppose, instead of pairings, we are provided with class labels for the training data in the two modalities. We cannot directly use KCCA in this case. One simple way to handle this situation is to generate data pairs using the class labels and then use KCCA with the generated pairs. We refer to this extension of KCCA as labeled KCCA in this paper.

Let $\{(x_i,y_i),\ i = 1,\hdots,N^x\}$ be $N^x$ labeled training samples from first modality with $y_i \in \{1,\hdots, P\}$ being the class label of feature $x_i$. Let $\{(z_j,w_j),\ j = 1,\hdots,N^z\}$ be $N^z$ labeled training samples from second modality with $w_j \in \{1,\hdots, P\}$ being the class label of feature $z_j$. Let $N^x_p$ and $N^z_p$ denote the number of training samples from class $p$ in first and second modalities respectively. Let $y = [y_1, \hdots, y_{N^x}]$ and $\vec{1}_p^x = \mathbbm{1}[y = p]$. Let $w = [w_1, \hdots, w_{N^z}]$ and $\vec{1}_p^z = \mathbbm{1}[w = p]$. 

In LKCCA, we form a training pair between samples $x_i$ and $z_j$ if $y_i = w_j$. A straightforward way to implement this is to replicate each data sample as many times as the number of samples from the same class in the other modality. But, this would unnecessarily increase the size of kernel matrices $K^x$ and $K^z$. Instead, LKCCA can be efficiently implemented without actually replicating the samples. This efficient implementation of LKCCA solves the ratio-trace problem~\eqref{eqn::ratio-trace-ker} with
\begin{equation}
K = K^x,\ \ \ L^{\prime} = EK^z (\sigma I + (1-\sigma)D^z K^z)^{-1}E^\top\ \ \text{and}\ \ L = D^x,
\end{equation}
where $D^x$ is a $N^x \times N^x$ diagonal matrix with $D^x_{i,i} = N^z_{y_i}$, $D^z$ is a $N^z \times N^z$ diagonal matrix with $D^z_{j,j} = N^x_{w_j}$, and $E = \sum_{p=1}^P \vec{1}_p^x \vec{1}_p^{z\top}$.

The latent space representations for samples $x \in \mathcal{R}^{d_1}$ and $z \in \mathcal{R}^{d_2}$ from first and second modalities respectively are given by  $x^{\prime} = \Gamma^\top [\mathcal{K}^x(x_1, x),\hdots, \mathcal{K}^x(x_{N^x}, x)]^\top$ and $z^{\prime} = \Xi^\top [\mathcal{K}^z(z_1, z), \hdots, \mathcal{K}^z(z_{N^z}, z)]^\top$, where 
\begin{equation}
\Xi = ((1-\sigma)D^zK^z + \sigma I)^{-1}E^\top K^x \Gamma\Lambda^{-\frac{1}{2}}.
\end{equation}
Here, $\Lambda$ is the diagonal matrix of non-zero generalized Eigenvalues given by~\eqref{eqn::gevd}. We refer the readers to supplementary material for further details about LKCCA.
\section{MKL-RT: MKL for Ratio-trace Problem}
In the MKL framework, the kernel function $\mathcal{K}$ is parametrized as a linear combination of pre-defined base kernel functions $\mathcal{K}^1, \hdots, \mathcal{K}^M$:
\begin{equation}
\mathcal{K} = \sum_{m=1}^M \mu_m \mathcal{K}^m,\ \mu_m \geq 0,\ \sum_{m=1}^M \mu_m = 1,
\end{equation}
and the weights $\mu = [\mu_1, \hdots, \mu_M]$ are learned from the data. Under this framework, MKL for ratio-trace problem can be formulated as the following optimization problem:
\begin{equation}
\begin{aligned}
& \underset{\Gamma,\ K,\ \mu}{\text{maximize}}\ \ \ \text{trace}\left[\left(\Gamma^\top ((1-\sigma)K L K + \sigma K)\Gamma\right)^{-1}\left(\Gamma^\top K L^{\prime} K \Gamma\right)\right]\\
& \hspace{47pt} K = \sum_{m=1}^M \mu_m K^m,\ \ \sum_{m=1}^M \mu_m = 1,\ \ \mu_m \geq 0\ \forall m,
\label{eqn::MKLorig}
\end{aligned}
\end{equation}
where $K^m$ is the kernel matrix corresponding to the base kernel function $\mathcal{K}^m$.

The optimization problem~\eqref{eqn::MKLorig} is a non-convex optimization problem.\ Nevertheless, the optimal $\mu^*$ for~\eqref{eqn::MKLorig} can be obtained by solving a different convex optimization problem as stated in the following theorem.\\[10pt]
\textbf{Theorem 1:} Let $L$ and $L^{\prime}$ be two symmetric positive semi-definite matrices with ranks $l$ and $l^{\prime}$ respectively. Let $\{(\alpha_i, u_i)\}_{i=1}^{l}$ and $\{(\beta_i, v_i)\}_{i=1}^{l^{\prime}}$ be the non-zero Eigenvalue-Eigenvector pairs of $L$ and $L^{\prime}$ respectively. Let $G = [\sqrt{\alpha_1}u_1,\hdots,\sqrt{\alpha_l}u_l]$ and $h_i = \sqrt{\beta_i}v_i$ for $i = 1,2,\hdots, l^{\prime}$. For $m = 1, \hdots, M$, let
\begin{equation}
\begin{aligned}
& S_m(\eta) = \frac{1}{\sigma}\sum_{i=1}^{l^{\prime}}  \left(\frac{\eta_i^\top \eta_i}{4(1-\sigma)} + \frac{\eta_i^\top G^\top K^m G \eta_i}{4 \sigma} - \eta_i^\top G^\top K^m h_i\right) + \text{trace}(K^m L^{\prime})
\end{aligned}
\label{eqn::Sm}
\end{equation}
be $M$ functions defined $\forall \eta = [\eta_1, \hdots, \eta_{l^{\prime}}] \in \mathcal{R}^{l \times l^{\prime}}.$ Let $\mu^{*}$ be optimal for the following convex optimization problem:
\begin{equation}
\begin{aligned}
& \underset{\zeta,\ \mu}{\text{maximize}}\ \ \zeta\\
& \text{subject to}\ \sum_{m = 1}^M\mu_m S_m(\eta) \geq \zeta,\ \ \forall \eta \in \mathcal{R}^{l \times l^{\prime}},\\
& \hspace{40pt} \sum_{m=1}^M \mu_m = 1,\ \ \mu_m \geq 0\ \forall m.
\label{eqn::SILP}
\end{aligned}
\end{equation}
Then $\mu^{*}$ is optimal for the optimization problem~\eqref{eqn::MKLorig}.\\[5pt]
\textbf{Proof:} Please refer to the supplementary material for the proof.\\[-5pt]

Note that the optimization problem~\eqref{eqn::SILP} is a semi-infinite linear program (SILP). Following~\cite{Sonnenburg06,Ye08}, we use an iterative approach to solve~\eqref{eqn::SILP}. In each iteration, the optimal $\mu$ and $\zeta$ are computed for a restricted subset of constraints $C$ in~\eqref{eqn::SILP}. Constraints that are not satisfied by current $\mu$ and $\zeta$ are added successively to the restricted problem until all the constraints are satisfied. For faster convergence, in each iteration, we add the constraint that maximizes the violation for current $\mu$ and $\zeta$. To find the maximum violating constraint, we solve
\begin{equation}
\eta^{*} = \underset{\eta}{\text{argmin}}\ \left(\sum_{m=1}^M \mu_m S_m(\eta) - \zeta\right).
\label{eqn::etacomp}
\end{equation}
Using the definition of $S_m(\eta)$ from equation~\eqref{eqn::Sm}, it can be easily verified that the optimum $\eta^{*} = [\eta_1^*, \eta_2^*,\hdots,\eta^*_{l^{\prime}}]$ for~\eqref{eqn::etacomp} can be obtained by individually solving for each $\eta_i^*$ using
\begin{equation}
\eta_i^{*} = \underset{\eta_i}{\text{argmin}} \left(\frac{\eta_i^\top \eta_i}{4\sigma(1-\sigma)} + \frac{\eta_i^\top G^\top K G \eta_i}{4 \sigma^2} - \frac{\eta_i^\top G^\top K h_i}{\sigma}\right),
\label{eqn::indetacomp}
\end{equation}
where $K = \sum_{m=1}^M \mu_m K^m$. Note that the optimization problem~\eqref{eqn::indetacomp} is an unconstrained quadratic program whose solution can be obtained by solving the following system of linear equations:
\begin{equation}
\left(\frac{I}{2(1-\sigma)} + \frac{G^\top K G}{2\sigma} \right)\eta_i^* = G^\top K h_i.
\label{eqn::linsys}
\end{equation}
\begin{table}[t]
\center
\caption{Algorithm for solving SILP~\eqref{eqn::SILP}.}
\vspace{3pt}
\begin{tabular}{|m{10.2cm}|}\hline
\vspace{1pt}
\hspace{3pt}\textbf{Input:} $L^{\prime},\ G,\ \{h_i\}_{i=1}^{l^{\prime}},\ \{K^m\}_{m=1}^M,\ T,\ \sigma,\ \epsilon$.\vspace{2pt}\\\hline
\vspace{2pt}
\hspace{3pt}\textbf{Initialization:} $\mu_m = \frac{1}{M}\ \forall m ,\ \zeta = +\infty, t = 1, C = \emptyset.$\vspace{2pt}\\\hline
\vspace{2pt}
\hspace{3pt}\textbf{while}\ \ $t\ \leq\ T$\\
\hspace{20pt} $K = \sum_{m=1}^M \mu_m K^m$\\
\hspace{20pt} \textbf{for} $i = 1,\hdots, l^{\prime}$\\
\hspace{37pt} Compute $\eta_i^*$ by solving $\left(\frac{I}{2(1-\sigma)} + \frac{G^\top K G}{2\sigma} \right)\eta_i^* = G^\top K h_i.$\\
\hspace{20pt} \textbf{end}\\
\hspace{20pt} \textbf{if}\ \ $\left \lvert 1 - \frac{\sum_{m=1}^M \mu_m S_m(\eta^*)}{\zeta} \right \rvert < \epsilon$ \hspace{5pt} \textbf{break};\\
\hspace{20pt} \textbf{else}\\
\hspace{37pt} Add $\eta^*$ to the constraint set $C$.\ \ Update $\mu$ and $\zeta$ by solving restricted\\
\hspace{37pt} version of~\eqref{eqn::SILP} using only $\eta \in C$.\\
\hspace{20pt} \textbf{end}\\
\hspace{20pt} t = t + 1;\\
\hspace{3pt}\textbf{end}\vspace{2pt}\\\hline
\vspace{2pt}
\hspace{3pt}\textbf{Output:} $\mu^*$ and $\zeta^*$.\vspace{2pt}\\\hline
\end{tabular}
\label{SILPalgo}
\end{table}
\indent Hence, in each iteration we solve $l^{\prime}$ linear systems to find the maximum violating constraint and one linear program to update $\mu$ and $\zeta$. Following~\cite{Sonnenburg06,Ye08}, we use the following stopping criterion:
\begin{equation}
\left \lvert 1 - \frac{\sum_{m=1}^M \mu_m S_m(\eta^*)}{\zeta} \right \rvert < \epsilon.
\end{equation}
Table~\ref{SILPalgo} summarizes the algorithm used for solving~\eqref{eqn::SILP}.\ This iterative algorithm is referred to as column generation technique and is guaranteed to converge~\cite{Sonnenburg06,Ye08}. 

Once the optimal $\mu^*$ is obtained, we can solve the ratio-trace problem~\eqref{eqn::ratio-trace-ker} using GEVD with $K^* = \sum_{m=1}^M \mu_m^*K^m$ to get the optimal $\Gamma^*$. Once $\mu^*$ and $\Gamma^*$ are known, the new non-linearly transformed representation for a data sample $x \in \mathcal{R}^d$ can be computed using $x^{\prime} = \Gamma^{*\top}[\mathcal{K}^*(x_1, x), \hdots, \mathcal{K}^*(x_N, x)]^\top$, where $\mathcal{K}^*(x_i, x) = \sum_{m=1}^M \mu_m^*\mathcal{K}^m(x_i, x)$. Table~\ref{MKLrtalgo} summarizes the proposed MKL-RT algorithm. 
\section{Experimental Evaluation}
\label{sec::Experiments}
\begin{table}[t]
\center
\vspace{-5pt}
\caption{MKL-RT algorithm.}
\vspace{3pt}
\begin{tabular}{|m{12cm}|}\hline
\vspace{3pt}
\hspace{3pt}\textbf{Input:} $L^{\prime},\ L,\ \{\mathcal{K}^m\}_{m=1}^M, T,\ \sigma,\ \epsilon$.\vspace{3pt}\\
\hline
\vspace{3pt}
\hspace{3pt}Compute $G = [\sqrt{\alpha_1}u_1,\hdots,\sqrt{\alpha_l}u_l],$ and $h_i = \sqrt{\beta_i}v_i$ for $i = 1,\hdots, l^{\prime}$, where $\{(\alpha_i, u_i)\}_{i=1}^{l}$\\
\hspace{3pt}and $\{(\beta_i, v_i)\}_{i=1}^{l^{\prime}}$ are the non-zero Eigenvalue-Eigenvector pairs of $L$ and $L^{\prime}$ respectively.\vspace{3pt}\\\hline
\vspace{3pt}
\hspace{3pt}Solve the SILP~\eqref{eqn::SILP} to obtain $\mu^*$ using the algorithm summarized in table~\ref{SILPalgo}.\vspace{3pt}\\\hline
\vspace{3pt}
\hspace{3pt}Solve the ratio-trace problem~\eqref{eqn::ratio-trace-ker} using GEVD with $K^* = \sum_{m=1}^M \mu_m^*K^m$ to get optimal $\Gamma^*$.\vspace{3pt}\\\hline
\vspace{3pt}
\hspace{3pt}The new non-linearly transformed representation for a data sample $x \in \mathcal{R}^d$ can be computed\\\hspace{3pt}using $x^{\prime} = \Gamma^{*\top}[\mathcal{K}^*(x_1, x), \hdots, \mathcal{K}^*(x_N, x)]^\top$, where $\mathcal{K}^*(x_i, x) = \sum_{m=1}^M \mu_m^*\mathcal{K}^m(x_i, x)$.\vspace{3pt}\\\hline
\end{tabular}
\label{MKLrtalgo}
\end{table}
We evaluated the proposed convex MKL-RT approach using three different instances of the ratio-trace problem: KFDA, KCCA and LKCCA (explained in section~\ref{sec::Instances}), covering two different applications: discriminative dimensionality reduction (for classification) and cross-modal retrieval. We used Caltech101~\cite{FeiFei04} and Oxford flowers17~\cite{Nilsback06} datasets for discriminative dimensionality reduction experiments, and Wikipedia articles~\cite{Rasiwasia10} and PascalVOC 2007~\cite{Hwang12} datasets for cross-modal retrieval experiments.
\subsection{Comparative Methods}
In all the experiments, we compare the proposed MKL-RT approach with the following methods:
\begin{itemize}
\item \textbf{AK-RT (Average kernel):} We solve the ratio-trace problem~\eqref{eqn::ratio-trace-ker} using the arithmetic mean kernel $\mathcal{K}_A$ defined as $\mathcal{K}_A = \frac{1}{M}\sum_{m=1}^M \mathcal{K}^m$.\\[-10pt]
\item \textbf{PK-RT (Product kernel):} We solve the ratio-trace problem~\eqref{eqn::ratio-trace-ker} using the geometric mean kernel $\mathcal{K}_P$ defined as $\mathcal{K}_P = (\prod_{m=1}^M \mathcal{K}^m)^{1/M}$.\\[-10pt]
\item \textbf{BIK-RT (Best individual kernel):} We solve the ratio-trace problem~\eqref{eqn::ratio-trace-ker} using the best kernel among $\{\mathcal{K}^m\}_{m=1}^M$.\\[-10pt]
\item \textbf{Non-convex MKL-DR\footnote{We used the code obtained from the authors of~\cite{Lin11} through personal correspondence.}:} We use the trace-ratio based non-convex MKL approach proposed in~\cite{Lin11}.
\end{itemize}

In the case of Caltech101 and Oxford flowers17 datasets, we use SVM and nearest neighbor (NN) rule for classification after discriminative dimensionality reduction. Hence, for these datasets, we also compare the proposed MKL-RT approach with the following SVM and NN based approaches:
\begin{itemize}
\item \textbf{Kernel SVM approaches:} Average kernel SVM (AK-SVM), product kernel SVM (PK-SVM), best individual kernel SVM (BIK-SVM) and MKL-SVM.\\[-10pt]
\item \textbf{Kernel NN approaches (without dimensionality reduction):} Average kernel NN (AK-NN), product kernel NN (PK-NN) and best individual kernel NN (BIK-NN). We computed the distances from kernels using
\begin{equation}
d^2(x_i, x_j) = \mathcal{K}(x_i, x_i) + \mathcal{K}(x_j, x_j) - \mathcal{K}(x_i, x_j) - \mathcal{K}(x_j, x_i).
\end{equation}
\end{itemize}
\subsection{KFDA for Discriminative Dimensionality Reduction}
In these experiments, we first performed dimensionality reduction using KFDA and then used SVM and NN rule for classification in the lower dimensional space. We used two different datasets, namely Caltech101~\cite{FeiFei04} and Oxford flowers17~\cite{Nilsback06}. The number of KFDA dimensions was chosen to be $P-1$, where $P$ is the number of classes.\\[10pt]
\noindent\textbf{Caltech101~\cite{FeiFei04}} is a multiclass object recognition dataset with 101 object categories and 1 background category. The authors of~\cite{Gehler09} precomputed 39 different kernels for this dataset using various image features and the kernel matrices are available online\footnote{http://files.is.tue.mpg.de/pgehler/projects/iccv09/}. We used these 39 kernels in our experiments and followed the experimental setup used in~\cite{Gehler09}. For brevity we omit the details of the features and refer to~\cite{Gehler09}. We report the results using all 102 classes of Caltech101. We repeated the experiment 5 times using different training and test splits and report the average results. The performance measure used is the mean prediction rate per class. We performed experiments using 5, 10, 15, 20, 25 and 30 training images per class and up to 50 test images per class. The regularization parameter $\sigma$ was chosen based on cross-validation.

Table~\ref{caltechresults} shows the recognition rates of various approaches for this dataset. For AK-SVM, PK-SVM, BIK-SVM and MKL-SVM, we report the results from~\cite{Gehler09}, which were obtained using the same kernel matrices and splits. We can make the following key observations from these results:
\begin{itemize}
\item Simple kernel NN approaches produce very poor results.\\[-8pt]
\item Performing discriminative dimensionality reduction gives a huge improvement with NN classifier (around 25-30\%) and a moderate improvement with SVM classifier (around 2\%). This is expected since the dimensionality reduction step makes samples from same class to be close to each other and samples from different classes to be far away from each other.\\[-8pt]
\item Among various KFDA approaches, the proposed convex MKL-RT approach works best with both NN and SVM classifiers.\\[-8pt]
\item The non-convex MKL-DR approach of~\cite{Lin11} performs poorly compared to AK-RT, PK-RT and MKL-RT approaches. The standard deviation of the MKL-DR approach is very high when the number of training samples is low (around 4\% for 5 training samples per class and 2.5\% for 10 training samples per class). This shows that the non-convex MKL-DR is overfitting the training data.\\[-8pt]
\item The proposed MKL-RT approach performs better (around 2.5\% on average) than MKL-SVM.
\end{itemize}
\setlength{\textfloatsep}{10pt}
\captionsetup[table]{font=scriptsize, skip=0pt}
\captionsetup[figure]{font=scriptsize}
\begin{table}[t]
\scriptsize
\center
\caption{Average recognition rates for Caltech101 dataset.}
\vspace{3pt}
\renewcommand{\arraystretch}{1.2}
\begin{tabular}{ c | c | c | c | c | c | c |}\cline{2-7}
& \multicolumn{6}{ c| }{Number of training images per class} \\ \hline
\multicolumn{1}{ |c| }{Method} & 5 & 10 & 15 & 20 & 25 & 30\\\hline\hline
\multicolumn{1}{ |c| }{AK-NN} & 27.3 $\pm$ 0.5	& 32.6 $\pm$ 0.6 & 36.1 $\pm$ 0.9 &	 38.0 $\pm$ 1.1 & 39.9 $\pm$ 1.0 & 42.4 $\pm$ 1.5\\\hline
\multicolumn{1}{ |c| }{PK-NN} & 27.5 $\pm$ 0.7	& 32.8 $\pm$ 1.0 & 36.4 $\pm$ 1.1 & 38.3 $\pm$ 1.2 &	40.4 $\pm$ 1.1 & 42.9 $\pm$ 1.4\\\hline
\multicolumn{1}{ |c| }{BIK-NN} & \textbf{29.4} $\mathbf{\pm}$ \textbf{0.9} & \textbf{35.5} $\mathbf{\pm}$ \textbf{1.2} & \textbf{39.5} $\mathbf{\pm}$ \textbf{0.7} & \textbf{41.3} $\mathbf{\pm}$ \textbf{1.0} & \textbf{43.3} $\mathbf{\pm}$ \textbf{0.8} & \textbf{45.7} $\mathbf{\pm}$ \textbf{1.1}\\\hline\hline
\multicolumn{1}{ |c| }{AK-RT-KFDA + NN} & 46.0 $\pm$ 0.8 & 57.6 $\pm$ 0.6 & 64.2 $\pm$ 0.8 & 68.1 $\pm$ 1.2 & 70.8 $\pm$ 0.9 & 73.6 $\pm$ 1.1\\\hline
\multicolumn{1}{ |c| }{PK-RT-KFDA + NN} & 45.0 $\pm$ 0.6 & 56.1 $\pm$ 0.6 & 62.6 $\pm$ 0.6 & 66.7 $\pm$ 0.7 & 69.4 $\pm$ 0.9 & 72.4 $\pm$ 1.0\\\hline
\multicolumn{1}{ |c| }{BIK-RT-KFDA + NN} & \textbf{46.1} $\mathbf{\pm}$ \textbf{0.6} & 54.8 $\pm$ 0.1	 & 59.7 $\pm$ 0.5 & 63.0 $\pm$ 0.8 & 65.4 $\pm$ 0.7 & 67.8 $\pm$ 0.7\\\hline
\multicolumn{1}{ |c| }{MKL-RT-KFDA + NN} & 45.7 $\pm$ 0.6	& \textbf{58.6} $\mathbf{\pm}$ \textbf{0.4} & \textbf{65.3} $\mathbf{\pm}$ \textbf{0.8} & \textbf{69.5} $\mathbf{\pm}$ \textbf{0.7} & \textbf{72.1} $\mathbf{\pm}$ \textbf{0.4}& \textbf{74.6} $\mathbf{\pm}$ \textbf{0.7}\\\hline
\multicolumn{1}{ |c| }{MKL-DR-KFDA + NN} & 40.2 $\pm$ 4.0	 & 53.6 $\pm$ 2.5 & 61.7 $\pm$ 1.4 & 65.5 $\pm$ 0.8 & 68.6 $\pm$ 0.9 & 72.1 $\pm$ 1.2\\\hline\hline
\multicolumn{1}{ |c| }{AK-SVM} & 44.4 $\pm$ 0.6	& \textbf{55.7} $\mathbf{\pm}$ \textbf{0.5} & 62.2 $\pm$ 1.1 &	66.1 $\pm$ 1.0 & 68.9 $\pm$ 1.0 & 71.6 $\pm$ 1.5\\\hline
\multicolumn{1}{ |c| }{PK-SVM} & 43.6 $\pm$ 0.7	& 54.7 $\pm$ 0.5 & 61.3 $\pm$ 0.9 & 65.4 $\pm$ 0.8 &	68.3 $\pm$ 0.7 & 71.3 $\pm$ 1.4\\\hline
\multicolumn{1}{ |c| }{BIK-SVM} & \textbf{46.1} $\mathbf{\pm}$ \textbf{0.9} & 55.6 $\pm$ 0.5 & 61.0 $\pm$ 0.2 & 64.3 $\pm$ 0.9 & 66.9 $\pm$ 0.8 & 69.4 $\pm$ 0.4\\\hline
\multicolumn{1}{ |c| }{MKL-SVM} & 42.1 $\pm$ 1.2 & 55.1 $\pm$ 0.7 & \textbf{62.3} $\mathbf{\pm}$ \textbf{0.8} & \textbf{67.1} $\mathbf{\pm}$\textbf{0.9} & \textbf{70.5} $\mathbf{\pm}$ \textbf{0.8} & \textbf{73.7} $\mathbf{\pm}$ \textbf{0.7}\\\hline\hline
\multicolumn{1}{ |c| }{AK-RT-KFDA + SVM} & 46.1 $\pm$ 0.8 & 57.6 $\pm$ 0.6 & 64.2 $\pm$ 0.8 & 68.1 $\pm$ 1.0 & 70.8 $\pm$ 0.9 & 73.7 $\pm$ 1.0\\\hline
\multicolumn{1}{ |c| }{PK-RT-KFDA + SVM} & 45.0 $\pm$ 0.6 & 56.2 $\pm$ 0.6 & 62.6 $\pm$ 0.7 & 66.7 $\pm$ 0.8 & 69.4 $\pm$ 0.9 & 72.5 $\pm$ 1.1\\\hline
\multicolumn{1}{ |c| }{BIK-RT-KFDA + SVM} & 46.1 $\pm$ 0.6 & 54.8 $\pm$ 0.2 & 59.8 $\pm$ 0.4 & 63.0 $\pm$ 0.8 & 65.5 $\pm$ 0.6 & 67.8 $\pm$ 0.7\\\hline
\multicolumn{1}{ |c| }{MKL-RT-KFDA + SVM} & \textbf{46.3} $\mathbf{\pm}$ \textbf{0.9}	& \textbf{58.9} $\mathbf{\pm}$ \textbf{0.4} & \textbf{65.5} $\mathbf{\pm}$ \textbf{0.7} & \textbf{69.7} $\mathbf{\pm}$ \textbf{1.0} & \textbf{72.2} $\mathbf{\pm}$ \textbf{0.4}& \textbf{74.7} $\mathbf{\pm}$ \textbf{0.7}\\\hline
\multicolumn{1}{ |c| }{MKL-DR-KFDA + SVM} & 40.2 $\pm$ 4.0 & 53.6 $\pm$ 2.5 & 61.7 $\pm$ 1.4 & 65.5 $\pm$ 0.8 & 68.6 $\pm$ 0.9 & 72.1 $\pm$ 1.3\\\hline
\end{tabular}
\label{caltechresults}
\end{table}
\begin{table}[t]
\scriptsize
\center
\parbox{.3\linewidth}{
\vspace{-3pt}
\center
\caption{Number of kernels selected by MKL-RT-KFDA for Caltech101.}
\vspace{3pt}
\renewcommand{\arraystretch}{1.1}
\begin{tabular}{ c | c | c | c | c | c | c |}\cline{2-7}
& \multicolumn{6}{ c| }{Training images per class} \\ \hline
\multicolumn{1}{ |c| }{Split} & 5 & 10 & 15 & 20 & 25 & 30\\\hline\hline
\multicolumn{1}{ |c| }{1} & 9 & 9 & 10 & 11 & 11& 11\\\hline
\multicolumn{1}{ |c| }{2} & 7 & 9 &  8 &   9 &  13 & 12\\\hline
\multicolumn{1}{ |c| }{3} & 9 & 10 & 10 & 11 & 12 & 12\\\hline
\multicolumn{1}{ |c| }{4} & 10 & 11	& 12 & 12 & 13 & 14\\\hline
\multicolumn{1}{ |c| }{5} & 8 & 10 & 10 & 11 & 11 & 11\\\hline
\end{tabular}
\label{selectedkernels}
}
\hspace{.15cm}
\quad
\parbox{.3\linewidth}{ 
\center
\captionof{figure}{Weights of proposed convex MKL-RT-KFDA for Caltech101.}
\vspace{3pt}
\includegraphics[width = 3.6cm, height = 2.4cm]{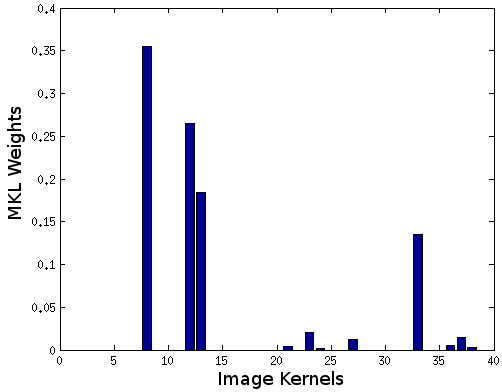}
\label{caltech_mkl_weights_convex}
}
\hspace{.15cm}
\quad
\parbox{.3\linewidth}{ 
\center
\captionof{figure}{Weights of non-convex MKL-DR-KFDA for Caltech101.}
\vspace{3pt}
\includegraphics[width = 3.6cm, height = 2.4cm]{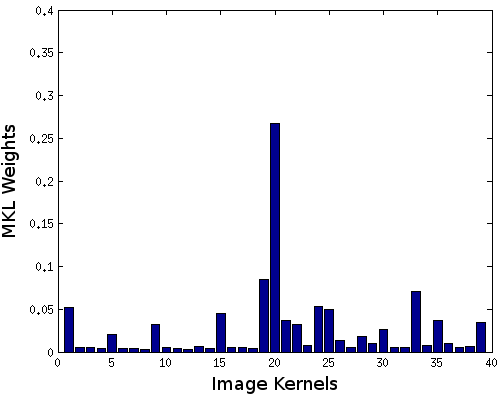}
\label{caltech_mkl_weights_nonconvex}
}
\end{table}

Table~\ref{selectedkernels} shows the number of kernels selected by MKL-RT-KFDA for the 5 splits used in our experiments.\ A kernel $\mathcal{K}^m$ is considered to be selected if its contribution is greater than 0.1\%, i.e., its coefficient $\mu_m$ is greater than $1/1000$.\ We can see that the number of kernels selected by MKL-RT-KFDA (around 7-14) is much less than the total number of kernels, which is 39 in this case. This clearly shows that the proposed approach can be successfully used to select features for discriminative dimensionality reduction. In contrast, the non-convex MKL-DR approach of~\cite{Lin11} ended up selecting all the 39 kernels (all the weights were greater than 0.001 after $\ell_1$ normalization). The main reason for this could be the lack of sparsity-promoting $\ell_1$-norm constraint (on the weights) in the MKL-DR formulation. Figures~\ref{caltech_mkl_weights_convex} and~\ref{caltech_mkl_weights_nonconvex} respectively show the kernel weights of MKL-RT-KFDA and MKL-DR-KFDA approaches for the fifth random split with 30 training samples per class.\\[10pt]
\noindent \textbf{Oxford flowers17~\cite{Nilsback06}} is a multiclass dataset consisting of 17 categories of flowers with 80 images per category. This dataset comes with 3 predefined splits into training ($17\times 40$ images), validation ($17\times 20$ images) and test ($17\times 20$ images) sets. Moreover, the authors of~\cite{Nilsback06} precomputed 7 distance matrices using various features and the matrices are available online\footnote{http://www.robots.ox.ac.uk/\texttildelow vgg/data/flowers/17/index.html}. For brevity we omit the details of the features and refer to~\cite{Nilsback06,Nilsback08}. We used these distance matrices and followed the same procedure as in~\cite{Gehler09} to compute 7 different kernels: $\mathcal{K}^m(x_i , x_j) = \text{exp}(-d_m(x_i , x_j)/\eta_m )$, where $\eta_m$ is the mean of the pairwise distances $d_m$ for the $m^{th}$ feature. We performed experiments using the three predefined splits and report the average results. The regularization parameter $\sigma$ was chosen based on cross-validation using the training and validation sets. 

Table~\ref{oxfordresults} shows the recognition rates of various approaches for this dataset. For the SVM-based approaches, we report the results from~\cite{Gehler09}, which were obtained using the same kernel matrices and splits. We can see that all the observations (except the high standard deviation of MKL-DR) made in the case of caltech101 hold true for oxford flowers17 dataset also. Figures~\ref{oxford_mkl_weights_convex} and~\ref{oxford_mkl_weights_nonconvex} respectively show the kernel weights for MKL-RT-KFDA and MKL-DR-KFDA approaches for the third split.
\begin{table}[t]
\scriptsize
\parbox{.4\linewidth}{
\center
\caption{Average recognition rates for Oxford flowers17 dataset.}
\vspace{3pt}
\renewcommand{\arraystretch}{1.2}
\begin{tabular}{| c | c | c | c |}\hline
\multicolumn{2}{ |c| }{NN} & \multicolumn{2}{ c| }{KFDA + NN}\\ \hline
AK &  71.9 $\pm$ 1.5 & AK-RT & 86.3 $\pm$ 1.5\\ \hline
PK &  \textbf{72.7} $\pm$ \textbf{2.1} & PK-RT & 86.7 $\pm$ 1.4\\ \hline
BIK  & 62.1 $\pm$ 1.9 & BIK-RT & 73.8 $\pm$ 2.0\\ \hline
\multicolumn{2}{ |c| }{} &  MKL-RT & \textbf{87.3} $\pm$ \textbf{1.5} \\\cline{3-4}
\multicolumn{2}{|c| }{}  & MKL-DR & 84.8 $\pm$ 1.1\\\hline\hline
\multicolumn{2}{ |c| }{SVM} & \multicolumn{2}{ c| }{KFDA + SVM}\\ \hline
AK & 84.9 $\pm$ 1.9  & AK-RT & 86.1 $\pm$ 1.7\\\hline
PK & \textbf{85.5} $\pm$ \textbf{1.2}  & PK-RT & 86.8 $\pm$ 1.6\\\hline
BIK  & 70.6 $\pm$ 1.6 & BIK-RT  & 74.1 $\pm$ 2.1\\\hline
MKL & 85.2 $\pm$ 1.5 & MKL-RT & \textbf{87.3} $\pm$ \textbf{1.7}\\\hline
\multicolumn{2}{ |c| }{}  & MKL-DR & 84.8 $\pm$ 1.1\\\hline
\end{tabular}
\label{oxfordresults}
}
\hspace{10pt}
\parbox{.25\linewidth}{
\vspace{19pt}
\center
\captionof{figure}{Weights of proposed convex MKL-RT-KFDA for Oxford flowers17.}
\vspace{3pt}
\includegraphics[width = 3cm, height = 3cm]{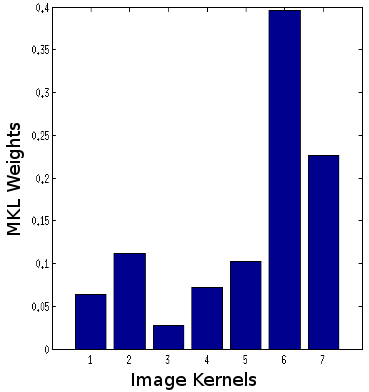}
\label{oxford_mkl_weights_convex}
}
\quad\quad
\parbox{.25\linewidth}{
\vspace{19pt}
\center
\captionof{figure}{Weights of non-convex MKL-DR-KFDA for Oxford flowers17.}
\vspace{3pt}
\includegraphics[width = 3cm, height = 3cm]{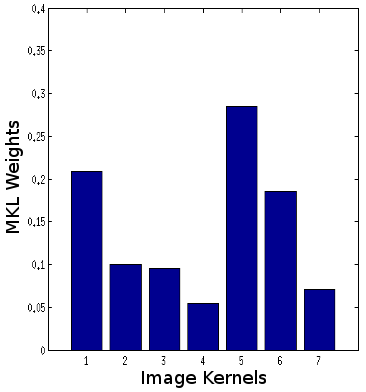}
\label{oxford_mkl_weights_nonconvex}
}
\end{table}

\subsection{KCCA for Cross-modal Retrieval}
In these experiments, we used KCCA to map the data from two different modalities (image and text) to a common latent space, and used the cosine distance in the latent space for cross-modal retrieval. We used Wikipedia articles~\cite{Rasiwasia10} dataset for these experiments. We measure the retrieval performance using mean average precision (MAP)~\cite{Rasiwasia10}.\\[10pt]
\textbf{Wikipedia articles~\cite{Rasiwasia10}} is a dataset of image-text pairs designed for cross-modal retrieval applications. It consists of 2173 training image-text pairs and 693 test image-text pairs which are grouped into 10 broad categories like art, history, etc. For text, we used a linear kernel generated from the 10-dimensional latent Dirichlet allocation model-based features provided by~\cite{Rasiwasia10}.\footnote{http://www.svcl.ucsd.edu/projects/crossmodal/} For images, we extracted various features and constructed 21 different kernels as described below:\\[3pt]
\indent\textbf{PHOG shape descriptor~\cite{Bosch07}:} The descriptor is a histogram of oriented ($Shp_{360}$) or unoriented ($Shp_{180}$) gradients computed on the output of a Canny edge detector. The $Shp_{360}$ histogram consists of 40 bins and the $Shp_{180}$ histogram consists of 20 bins.\ We generated 4 kernels corresponding to different levels of spatial pyramid~\cite{Lazebnik06} from both $Shp_{360}$ and $Shp_{180}$. Each kernel is an RBF kernel based on the $\chi^2$ distance between histograms.\\[3pt]
\indent\textbf{SIFT appearance descriptor:} We computed the grayscale SIFT~\cite{Lowe04} descriptors on a regular grid on the image with a spacing of 2 pixels and for four different sizes $r  = 4, 6, 8, 10.$ We followed two different approaches, namely BOW model with 1000 codewords and second-order pooling~\cite{Carreira12}, to obtain region descriptors from the SIFT descriptors. In each case, we generated 3 kernels corresponding to different levels of spatial pyramid. In the case of BOW model, each kernel is an RBF kernel based on the $\chi^2$ distance between histograms. In the case of second-order pooling, each kernel is an RBF kernel based on the log-Euclidean distance~\cite{Arsigny06} between covariance matrices.\\[3pt]
\indent\textbf{LBP texture features~\cite{Ojala02}:} We used the histograms of uniform rotation invariant $LBP_{8,1}$ features and generated 3 kernels corresponding to different levels of spatial pyramid. Each kernel is an RBF kernel based on the $\chi^2$ distance between histograms.\\[3pt]
\indent\textbf{Region covariance features~\cite{Tuzel06}:} We used the covariance of simple per-pixel features described in~\cite{Tuzel06}. We generated 3 kernels corresponding to different levels of spatial pyramid. Each kernel is an RBF kernel based on the Log-Euclidean distance~\cite{Arsigny06} between covariance matrices.\\[3pt]
\indent\textbf{GIST image descriptor~\cite{Oliva01}}: We generated an RBF kernel using the 512-dimensional GIST descriptor that records the pooled steerable filter responses within a grid of spatial cells across the image.

The number of KCCA dimensions was chosen to be 9 and the regularization parameter $\sigma$ was chosen using cross-validation. Table~\ref{wikipedia_results} shows the MAP scores of various KCCA approaches on this dataset for text and image queries. We can clearly see that the proposed MKL-RT approach gives the best retrieval performance. Similar to KFDA experiments, the MKL-DR approach performs poorly compared to AK-RT, PK-RT and MKL-RT. Figures~\ref{wikipedia_mkl_weights_convex} and~\ref{wikipedia_mkl_weights_nonconvex} respectively show the kernel weights for MKL-RT-KCCA and MKL-DR-KCCA approaches. For this dataset, the proposed convex MKL-RT-KCCA approach selected 9 kernels out of 21, whereas the non-convex MKL-DR-KCCA approach ended up selecting all the kernels. This clearly shows that the proposed approach can be successfully used for feature selection in cross-modal retrieval applications.
\setlength{\tabcolsep}{1.5pt}
\begin{table}[t]
\scriptsize
\parbox{.4\linewidth}{
\center
\vspace{-6pt}
\caption{MAP scores for Wikipedia dataset.}
\vspace{3pt}
\renewcommand{\arraystretch}{1.2}
\begin{tabular}{ c | c | c | c } \cline{2-3}
& \multicolumn{2}{c |}{Query} & \\\hline
\multicolumn{1}{|c|}{Method} & Text & Image & \multicolumn{1}{c|}{Average}\\\hline
\multicolumn{1}{|c|}{AK-RT-KCCA} & 27.74 & 31.43 & \multicolumn{1}{c|}{29.58}\\\hline
\multicolumn{1}{|c|}{PK-RT-KCCA} & 27.86 & 31.49 & \multicolumn{1}{c|}{29.68}\\\hline
\multicolumn{1}{|c|}{BIK-RT-KCCA} & 26.45 & 30.4 & \multicolumn{1}{c|}{28.43}\\\hline
\multicolumn{1}{|c|}{MKL-RT-KCCA} & \textbf{28.14} & \textbf{32.73} & \multicolumn{1}{c|}{\textbf{30.43}}\\\hline
\multicolumn{1}{|c|}{MKL-DR-KCCA} & 27.44 & 30.74 & \multicolumn{1}{c|}{29.09}\\\hline
\end{tabular}
\label{wikipedia_results}
}
\parbox{.28\linewidth}{
\center
\captionof{figure}{Weights of proposed convex MKL-RT-KCCA for Wikipedia.}
\vspace{3pt}
\includegraphics[width = 3.3cm, height = 2.4cm]{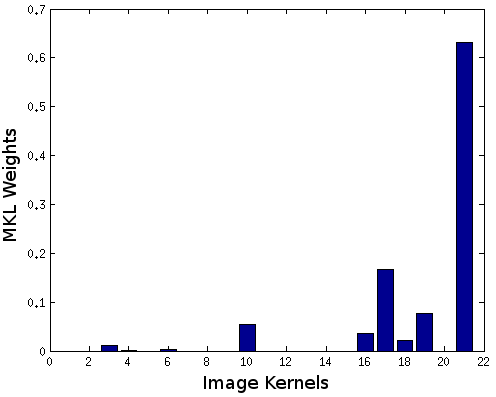}
\label{wikipedia_mkl_weights_convex}
}
\hspace{5pt}
\parbox{.29\linewidth}{
\center
\captionof{figure}{Weights of non-convex MKL-DR-KCCA for Wikipedia.}
\vspace{3pt}
\includegraphics[width = 3.3cm, height = 2.4cm]{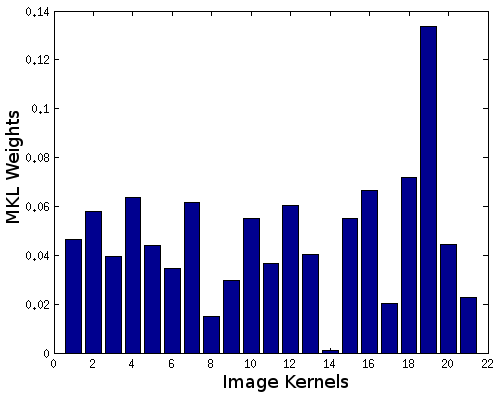}
\label{wikipedia_mkl_weights_nonconvex}
}
\end{table}
\subsection{LKCCA for Cross-modal Retrieval}
In these experiments, we used LKCCA to map the data from two different modalities (image and text) to a common latent space, and used the cosine distance in the latent space for cross-modal retrieval. We used Wikipedia articles~\cite{Rasiwasia10} and PascalVOC 2007~\cite{Hwang12} datasets for these experiments. The number of LKCCA dimensions was chosen to be $P-1$, where $P$ is the number of classes. We use the mean average precision to measure the retrieval performance.

For the Wikipedia dataset, we used the same image and text kernels that were used in KCCA experiments. Table~\ref{wikipedia_lkcca_results} shows the MAP scores of various LKCCA approaches on this dataset for text and image queries. We can clearly see that the proposed MKL-RT approach gives the best retrieval performance. Similar to KFDA and KCCA experiments, the MKL-DR approach performs poorly compared to AK-RT, PK-RT and MKL-RT. Figures~\ref{wikipedia_lkcca_mkl_weights_convex} and~\ref{wikipedia_lkcca_mkl_weights_nonconvex} respectively show the kernel weights for MKL-RT-LKCCA and MKL-DR-LKCCA approaches. For this dataset, the proposed convex MKL-RT-LKCCA approach selected 12 kernels out of 21, whereas the non-convex MKL-DR-LKCCA approach selected 18 kernels.\\[10pt]
\begin{table}[t]
\scriptsize
\parbox{.4\linewidth}{
\center
\caption{MAP scores for Wikipedia dataset.}
\vspace{3pt}
\renewcommand{\arraystretch}{1.2}
\begin{tabular}{ c | c | c | c } \cline{2-3}
& \multicolumn{2}{c |}{Query} & \\\hline
\multicolumn{1}{|c|}{Method} & Text & Image & \multicolumn{1}{c|}{Average}\\\hline
\multicolumn{1}{|c|}{AK-RT-LKCCA} & 27.28 & 31.02 & \multicolumn{1}{c|}{29.15}\\\hline
\multicolumn{1}{|c|}{PK-RT-LKCCA} & 27.16 & 30.75 & \multicolumn{1}{c|}{28.96}\\\hline
\multicolumn{1}{|c|}{BIK-RT-LKCCA} & 26.43 & 30.45 & \multicolumn{1}{c|}{28.44}\\\hline
\multicolumn{1}{|c|}{MKL-RT-LKCCA} & \textbf{27.92} & \textbf{32.09} & \multicolumn{1}{c|}{\textbf{30.01}}\\\hline
\multicolumn{1}{|c|}{MKL-DR-LKCCA} & 26.73 & 29.1 & \multicolumn{1}{c|}{27.92}\\\hline
\end{tabular}
\label{wikipedia_lkcca_results}
}
\parbox{.28\linewidth}{
\center
\captionof{figure}{Weights of proposed convex MKL-RT-LKCCA for Wikipedia.}
\vspace{3pt}
\includegraphics[width = 3.3cm, height = 2.2cm]{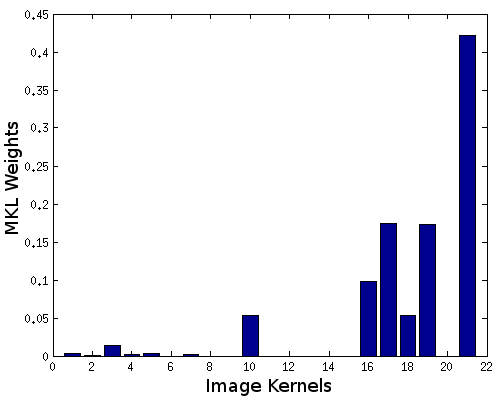}
\label{wikipedia_lkcca_mkl_weights_convex}
}
\hspace{5pt}
\parbox{.29\linewidth}{
\center
\captionof{figure}{Weights of non-convex MKL-DR-LKCCA for Wikipedia.}
\vspace{3pt}
\includegraphics[width = 3.3cm, height = 2.2cm]{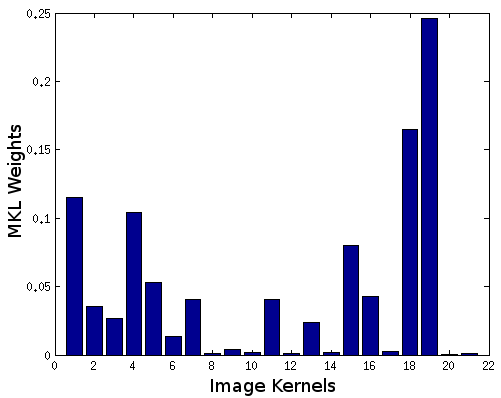}
\label{wikipedia_lkcca_mkl_weights_nonconvex}
}
\end{table}
\noindent\textbf{PascalVOC 2007~\cite{Hwang12}} dataset consists of 5011 training image-text pairs and 4952 test image-text pairs corresponding to 20 different object categories. Since some of the images are multi-labeled, following~\cite{Sharma12,Wang13}, we selected the images with only one object. This gave us 2799 training image-text pairs and 2820 test image-text pairs. For text, we used a linear kernel generated from the \textit{absolute} and \textit{relative} tag rank features provided by~\cite{Hwang12}.\footnote{http://vision.cs.utexas.edu/projects/tag/bmvc10.html} For images, we extracted various different features (same as those used for the Wikipedia dataset) and constructed 21 different kernels.

Table~\ref{pascal_results} shows the MAP scores of various LKCCA approaches on this dataset for text and image queries. For image query, the proposed MKL-RT approach gives the best retrieval performance and the MKL-DR approach performs very poorly. For text query, the MKL-DR approach gives the best performance and the proposed MKL-RT approach is the second best. Considering the average retrieval performance, the proposed MKL-RT approach is the best. Figures~\ref{pascal_mkl_weights_convex} and~\ref{pascal_mkl_weights_nonconvex} respectively show the kernel weights for MKL-RT-LKCCA and MKL-DR-LKCCA approaches. For this dataset, the proposed convex MKL-RT-LKCCA approach selected 8 kernels out of 21, whereas the non-convex MKL-DR-LKCCA approach selected 20 kernels.

For PascalVOC dataset, we also performed experiments using KCCA. The results produced by all the KCCA methods were much lower than the corresponding results of LKCCA methods. So, in the interest of space, we are not presenting those results.

\begin{table}[t]
\scriptsize
\parbox{.4\linewidth}{
\center
\caption{MAP scores for PascalVOC dataset}
\vspace{3pt}
\renewcommand{\arraystretch}{1.2}
\begin{tabular}{ c | c | c | c } \cline{2-3}
& \multicolumn{2}{c |}{Query} & \\\hline
\multicolumn{1}{|c|}{Method} & Text & Image & \multicolumn{1}{c|}{Average}\\\hline
\multicolumn{1}{|c|}{AK-RT-LKCCA} & 53.1 & 51.41 & \multicolumn{1}{c|}{52.26}\\\hline
\multicolumn{1}{|c|}{PK-RT-LKCCA} & 52.05 & 50.87 & \multicolumn{1}{c|}{51.46}\\\hline
\multicolumn{1}{|c|}{BIK-RT-LKCCA} & 51.81 & 51.39 & \multicolumn{1}{c|}{51.6}\\\hline
\multicolumn{1}{|c|}{MKL-RT-LKCCA} & 53.56 & \textbf{52.92} & \multicolumn{1}{c|}{\textbf{53.24}}\\\hline
\multicolumn{1}{|c|}{MKL-DR-LKCCA} & \textbf{55.05} & 46.49 & \multicolumn{1}{c|}{50.77}\\\hline
\end{tabular}
\label{pascal_results}
}
\parbox{.28\linewidth}{
\center
\captionof{figure}{Weights of proposed convex MKL-RT-LKCCA for PascalVOC.}
\vspace{3pt}
\includegraphics[width = 3.3cm, height = 2.2cm]{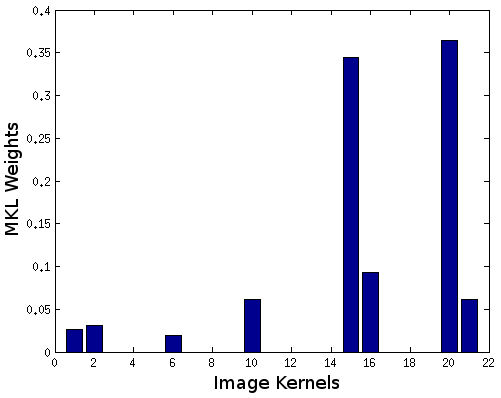}
\label{pascal_mkl_weights_convex}
}
\hspace{5pt}
\parbox{.29\linewidth}{
\center
\captionof{figure}{Weights of non-convex MKL-DR-LKCCA for PascalVOC.}
\vspace{3pt}
\includegraphics[width = 3.3cm, height = 2.2cm]{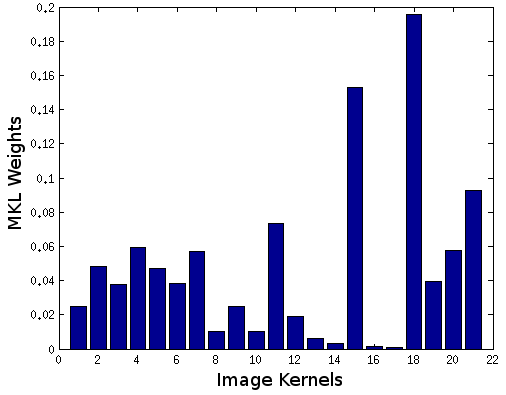}
\label{pascal_mkl_weights_nonconvex}
}
\end{table}
\section{Conclusion and Future Work}
In this paper, we showed that MKL can be formulated as a convex optimization problem for a large class of ratio-trace problems that includes many popular algorithms like LDA, SDA, SILDA, LDE, MFA, LPP, NPE, CCA and Orthonormal PLS-SB. We also provided an optimization procedure that is guaranteed to converge to the global optimum of the proposed convex optimization problem. We performed experiments using three different instances of the ratio-trace problem and demonstrated that the proposed MKL-RT approach can be successfully used to select features for discriminative dimensionality reduction and cross-modal retrieval. We also showed that the proposed convex MKL-RT approach performs better than the non-convex MKL-DR approach of~\cite{Lin11}.

In the near future, we plan to test our approach on various other instances of the ratio-trace problem. Similar to the lines of $\ell_p$-MKL-SVM and $\ell_p$-MKL-LDA, we also plan to extend this work to $\ell_p$-MKL-RT.

\bibliographystyle{splncs03}
\bibliography{eccv2014submission}
\end{document}